\documentclass[letterpaper, 10 pt, journal, twoside]{IEEEtran}

\usepackage[font=small]{caption}
\usepackage{amsmath}
\usepackage{balance}
\usepackage[ruled,linesnumbered]{algorithm2e}
\usepackage{graphicx}
\usepackage{subfigure}

\usepackage{enumitem}
\usepackage{amssymb}
\usepackage{multirow}
\usepackage{hyperref}
\usepackage{float}
\hypersetup{hidelinks}
\usepackage{tabularx}
\usepackage{wrapfig}
\usepackage{color}
\newcommand{\rj}[1]{{\color{black}#1}}
\newcolumntype{C}[1]{>{\centering\let\newline\\\arraybackslash\hspace{0pt}}m{#1}}
\ifCLASSINFOpdf
\else
\fi
\begin{document}
%

%
%
%

\title{\LARGE \bf
	Deep Samplable Observation Model \\ for Global Localization and Kidnapping 
}

\author{Runjian Chen$^{1}$, Huan Yin$^{1}$, Yanmei Jiao$^{1}$, Gamini Dissanayake$^{2}$, Yue Wang$^{1}$ and Rong Xiong$^{1}$
	\thanks{Manuscript received Oct 15th, 2020; Revised Jan 6th, 2021; Accepted Feb 4th, 2021.}
	\thanks{This paper was recommended for publication by Editor Javier Civera upon evaluation of the Associate Editor and Reviewers' comments.
		} 
	\thanks{$^{1}$Runjian Chen, Yue Wang, Huan Yin, Yanmei Jiao and Rong Xiong are the State Key Laboratory of Industrial Control Technology and Institute of Cyber-Systems and Control, Zhejiang University, Zhejiang, China. Yue Wang is the corresponding author {\tt\small wangyue@iipc.zju.edu.cn}.}%
	\thanks{$^{2}$Gamini Dissanayake is with the Center for Autonomous System, University of Technology Sydney.}%
}

%
%

\markboth{IEEE Robotics and Automation Letters. Preprint Version. Accepted Feb, 2021}
{Chen \MakeLowercase{\textit{et al.}}: DSOM} 

%



\maketitle

\begin{abstract}
Global localization and kidnapping are two challenging problems in robot localization. The popular method, Monte Carlo Localization (MCL) addresses the problem by iteratively updating a set of particles with a ``sampling-weighting'' loop. Sampling is decisive to the performance of MCL \cite{thrun2001robust}. However, traditional MCL can only sample from a uniform distribution over the state space. Although variants of MCL propose different sampling models, they fail to provide an accurate distribution or generalize across scenes. To better deal with these problems, we present a distribution proposal model named Deep Samplable Observation Model (DSOM). DSOM takes a map and a 2D laser scan as inputs and outputs a conditional multimodal probability distribution of the pose, making the samples more focusing on the regions with higher likelihood. With such samples, the convergence is expected to be more effective and efficient. Considering that the learning-based sampling model may fail to capture the accurate pose sometimes, we furthermore propose the Adaptive Mixture MCL (AdaM MCL), which deploys a trusty mechanism to adaptively select updating mode for each particle to tolerate this situation. Equipped with DSOM, AdaM MCL can achieve more accurate estimation, faster convergence and better scalability than previous methods in both synthetic and real scenes. Even in real environments with long-term changes, AdaM MCL is able to localize the robot using DSOM trained only by simulation observations from a SLAM map or a blueprint map. Source code for this paper is available \href{https://github.com/Runjian-Chen/AdaM_MCL}{\texttt{here}}.
\end{abstract}


%
\IEEEpeerreviewmaketitle

\section{Introduction}
\IEEEPARstart{T}{he} ability to accurately localize a robot is the fundamental requirement for many autonomous tasks, including motion planning, decision making and control \cite{cox1991blanche,fukuda1993navigation,hinkel1989environment,leonard1992dynamic}. In this paper, we focus on the global localization and kidnapping problem on 2D scenes with 2D laser observation.  Given real-time motion information and 2D laser observation, global localization algorithms aim to estimate the pose of the robot in a map of the environment without any prior about the robot pose. For kidnapping, a more complicated problem where the robot is suddenly taken to some other place without being told, the algorithm should be able to detect this situation and recover from it.

Monte Carlo Localization (MCL) \cite{thrun2001robust} is arguably the most popular and efficient algorithm \cite{Stachniss2014} for these two problems. MCL uses a set of particles with weights to represent the estimated probability distribution of the robot pose and iteratively deploys a ``sampling-weighting'' routine to update this set. The essential component of MCL is \textbf{sampling} particles from  a \textbf{proposal distribution} \cite{thrun2001robust}.

\setlength{\textfloatsep}{0.8pt}

\begin{figure}
	\flushleft
	\subfigure[\textbf{Traditional MCL}: only uniform distribution.]{
		\begin{minipage}[t]{0.96\linewidth}
			\flushleft
			\includegraphics[width=3.3in]{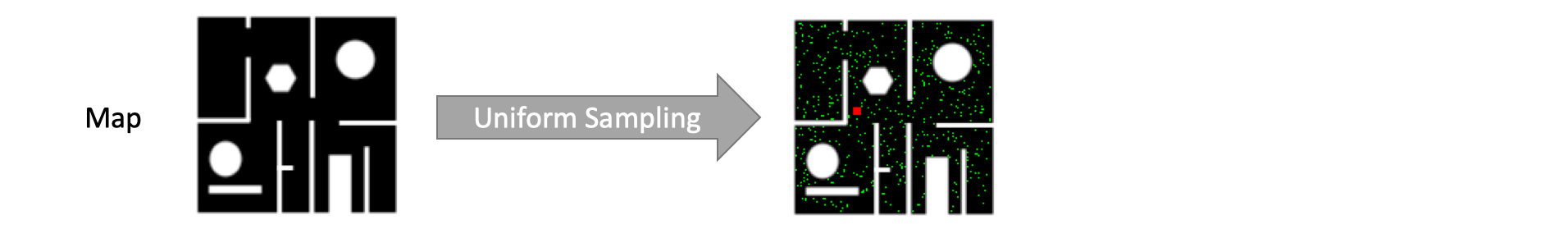}
		\end{minipage}%
		
	}%
	\vspace{-0.2cm}
	\flushleft
	\subfigure[\textbf{Dual MCL}: sample from limited poses with handcrafted features.]{
		\begin{minipage}[t]{0.96\linewidth}
			\flushleft
			\includegraphics[width=3.3in]{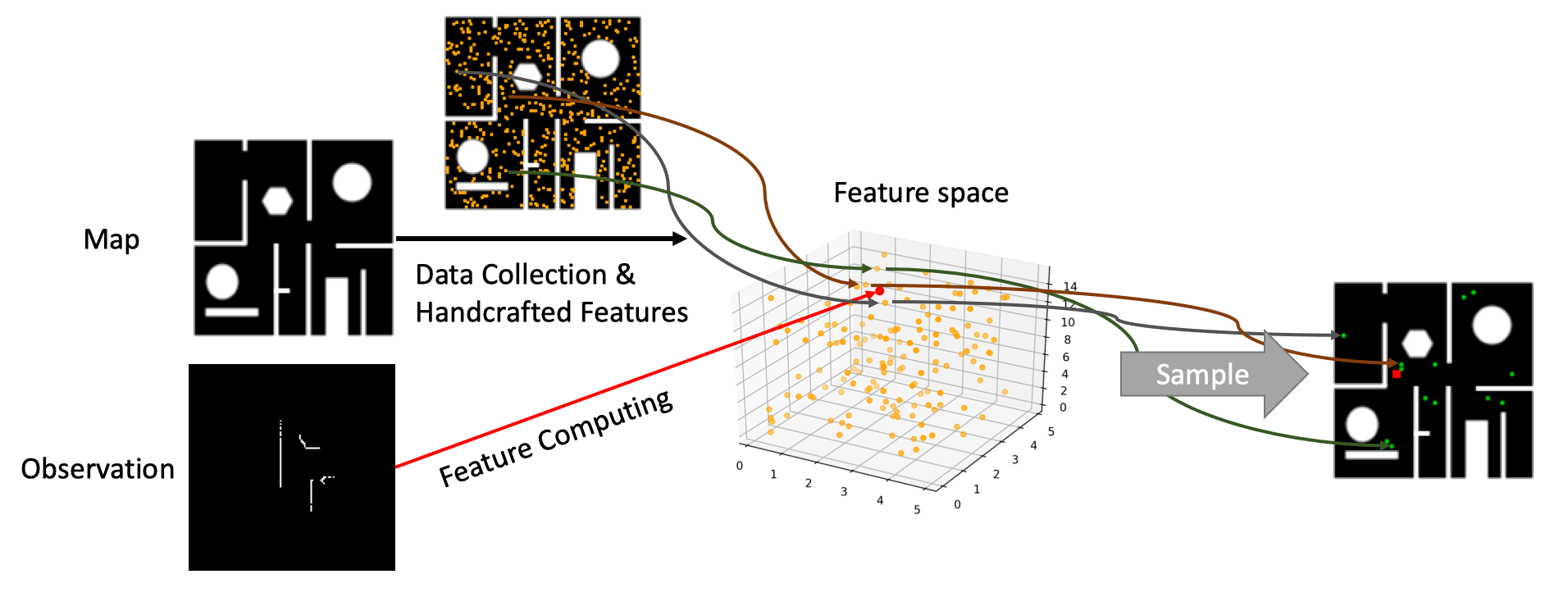}
		\end{minipage}%
		
	}%
	\vspace{-0.2cm}
	\flushleft
	\subfigure[\textbf{DSOM}: distribution over the entire space.]{
		\begin{minipage}[t]{0.96\linewidth}
			\flushleft
			\includegraphics[width=3.3in]{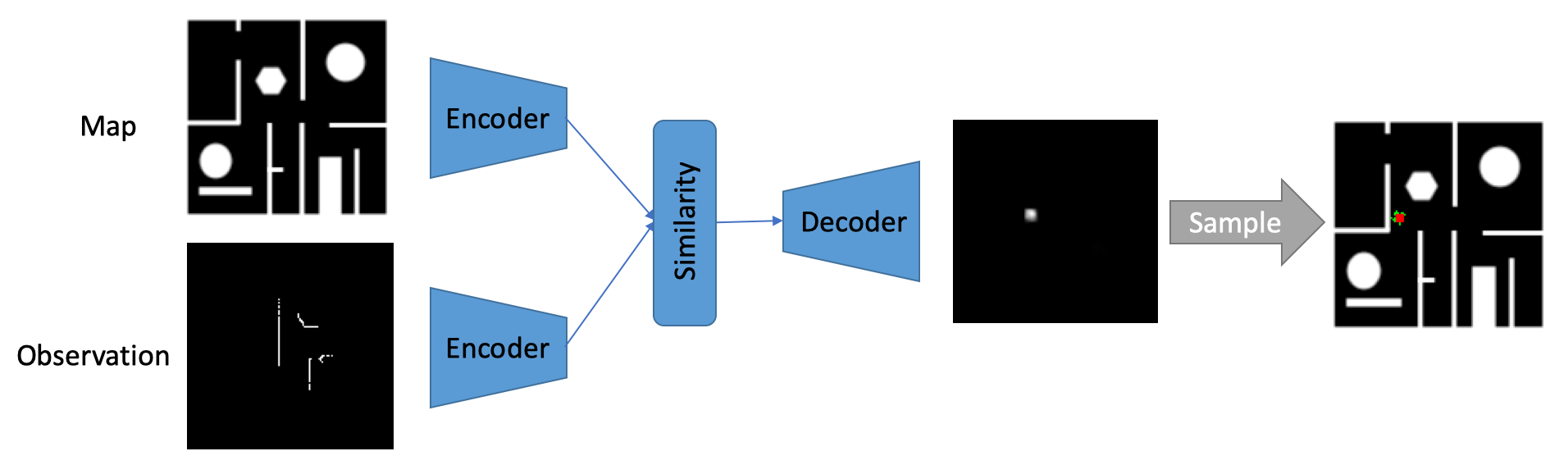}
		\end{minipage}%
		
	}%
	\vspace{-0.2cm}
	\caption{Sampling models in MCL, Dual MCL and AdaM MCL.}
	\label{fig:intro}
	\vspace{0.15cm}
\end{figure}


In global localization and kidnapping problems, traditional MCL can only sample from a uniform distribution over the entire state space. On the one hand, if the particle number is small, there might not be any particle around the accurate pose, resulting in an inaccurate estimation of the robot pose. On the other hand, if the particle number is large enough to cover the whole state space, 
the computational cost is too high and makes it impossible to finish the update in real-time. To make sampling more effective and efficient, Dual MCL in \cite{thrun2001robust} proposes to sample from observation with three handcrafted features, as shown in Figure \ref{fig:intro}. Mixture MCL in \cite{thrun2001robust}, which combines traditional MCL and Dual MCL by executing the former with probability $p$ or the latter with probability $1-p$ in an updating iteration, achieves the best performance among variants of MCL \cite{thrun2001robust,nummiaro2003adaptive,okuma2004boosted}. However, these handcrafted features are far from enough to extract meaningful information from observation and it relies on data collection in the same environment where it is deployed, making it unrobust and unable to generalize across scenes.

Recently, many researchers focus on deep learning techniques for providing a proposal distribution \cite{kim20191,xu2019robust,angelina2018pointnetvlad,jonschkowski2018differentiable,yin2018locnet}. However, the proposal distribution generation in \cite{jonschkowski2018differentiable} only takes the observation as input and it requires training data collection in the testing environment, which makes it unable to generalize across scenes. \cite{kim20191,xu2019robust,angelina2018pointnetvlad,yin2018locnet} are only able to provide a proposal distribution with a fixed number of probability peaks, which limits their performance because one observation can usually correspond to various numbers of poses where the robot can get similar observations.

To address the problems mentioned above, we propose the Deep Samplable Observation Model (DSOM), which utilizes the feature extraction and generalization ability of the convolutional neural network to efficiently generate a probability map over the entire state space with an adaptive number of probability peaks. Figure \ref{fig:intro} describes the difference between sampling models in MCL, Mixture MCL and DSOM. Although the output proposal distribution from DSOM can easily be incorporated in the Mixture MCL framework, the performance is unstable due to the combining mechanism in the original Mixture MCL framework. Thus we further propose Adaptive Mixture MCL (AdaM MCL), which introduces a trusty mechanism to divide the particle set into two parts: highly trusted one and untrusted one, and deploy traditional MCL and dual MCL respectively for these two parts. \rj{AdaM MCL is able to generalize across scenes and localize the robot with high accuracy in global localization and kidnapping problems.}

\section{Related Work}
Global localization and kidnapping have long been traditional topics in robotics. From traditional methods built upon probabilistic framework to recent deep learning techniques, we divide algorithms on this topic into the following three categories: (1) traditional techniques (2) deep-sampling-model-aided particle filter (3) differentiable particle filter.\vspace{0.1cm}

\noindent\textbf{Traditional Techniques. } Based on the traditional probabilistic framework, this category is embraced by extended Kalman Filter (EKF)  \cite{welch1995introduction,julier1997new}  and Monte Carlo Localization (MCL) \cite{thrun2001robust,Fox-1999-14948}. EKF depends on a unimodal assumption and is restricted to a specific distribution, i.e. the Gaussian distribution. Thus it is only able to handle the pose tracking problem where the prior about the robot pose (eg. mean and variance) is required. Meanwhile, MCL, also called the particle filter, utilizes a set of particles to represent the multimodal distribution and eliminate the constraint brought by a specific distribution, achieving relatively satisfying results on global localization and kidnapping problems. Among variants of MCL, Mixture MCL \cite{thrun2001robust} is the most successful one. It introduces Dual MCL, which utilizes three handcrafted features (the location of a sensor scan's center of gravity and the average distance measurement) for better sampling particles from observation, as shown in Figure \ref{fig:intro}. Mixture MCL combines traditional MCL and Dual MCL for better performance. However, the handcrafted features are not informative and data collection in the same scene is required, which is always inapplicable in the real world. Thus Mixture MCL suffers from the lack of generalization ability and unrobustness, especially in highly symmetrical environments.\vspace{0.1cm}

\noindent\textbf{Deep Sampling Model for Aiding MCL. } Works including \cite{kim20191,xu2019robust,angelina2018pointnetvlad,kendall2015posenet,yin20193d} deploy the deep neural network to provide a proposal distribution, which boosts the performance of MCL. They can be classified into two categories by how many probability peaks they provide. The first one is \textbf{unimodal}. Represented by \cite{kim20191,kendall2015posenet}, a classifier or a regression network is trained to find the pose with the largest possibility to obtain the observation in a given map, which provides only a unimodal distribution. However, as the environment can be very symmetrical and dynamic, observations at different poses can be very similar and one observation usually corresponds to several poses, for which the unimodal distribution might quickly lose track of the robot pose. The second category is \textbf{Top N}. Embraced by \rj{\cite{xu2019robust,angelina2018pointnetvlad,yin2018locnet,yin20193d}}, this kind of methods train deep models to extract features of the observation and features of templates in the database, with which they are able to propose top $N$ possible regions. This category has two limitations. First, it can only choose the top $N$ regions, where $N$ is a fixed handcrafted parameter. However, in real scenes, one observation might correspond to different numbers of poses and $N$ limits this model's flexibility to approximate the accurate multimodal distribution. Secondly, the output of this kind of model is simply $N$ regions and additional effort should be taken to construct a probability distribution over the state space. \vspace{0.1cm}

\noindent\textbf{Differentiable Particle Filter. } \cite{jonschkowski2018differentiable,ma2019particle,karkus2018particle} try to make the whole updating process of particle filter differentiable. However, the generation process of proposal distribution in \cite{jonschkowski2018differentiable} does not take the map as inputs and it is trained on the exact map where they collect testing sequences, making it unable to generalize across different scenes. Also, in \rj{\cite{ma2019particle,karkus2018particle}}, the methods do not consider sampling from a deep proposal and are still only able to sample from a uniform distribution at the very beginning. Thus it is doubtful whether these methods \rj{are} able to cope with kidnapping and there are no experiments on kidnapping in these works.

To briefly summarize, a good sampling model should provide an accurate multimodal probability distribution with an adaptive number of probability peaks. Additionally, generalization ability, low time consumption and scalability over scene sizes are also important requirements. Moreover, as training samples for the sampling model may not cover every possible situation, the predicted probability distribution will occasionally fail to capture the correct pose. The simple combination of traditional MCL and Dual MCL in Mixture MCL may result in losing track of the correct pose if Mixture MCL barely uses the Dual MCL branch when its sampling model fails to capture the correct pose. In this paper, we focus on these two challenges and propose the Deep Sample Observation Model (DSOM) and Adaptive Mixture MCL (AdaM MCL) to address them.

\begin{figure}[h]
	\begin{center}
		\includegraphics[height=0.58\linewidth]{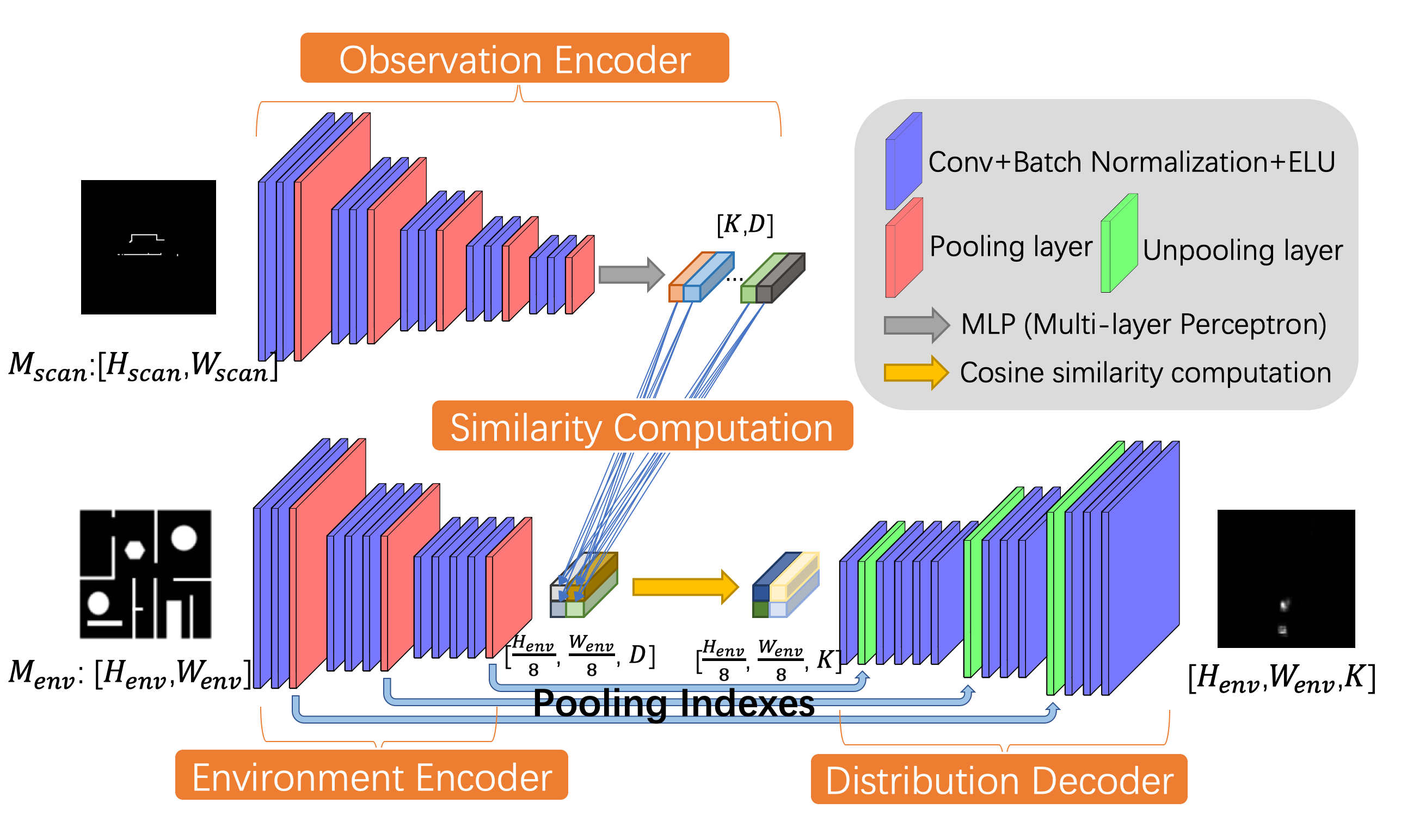} %
	\end{center}
	\vspace{-0.3cm}
	\caption{The architecture of Deep Samplable Observation Model.}
	
	\label{fig:DMOM}
	\vspace{0.2cm}
\end{figure}

\section{Deep Samplable Obervation Model}
\label{sec: Deep Multimodal Obervation Model}
In this section, we propose our Deep Samplable Observation Model (DSOM). As shown in Figure \ref{fig:DMOM}, the inputs of DSOM are two images, one for the environment and one for the observation information (here we plot the 2D laser ranges on an image), represented by two matrices $\mathbf{M}_{env}\in \mathbb{R}^{H_{env}\times W_{env}}$ and $\mathbf{M}_{scan}\in \mathbb{R}^{H_{scan}\times W_{scan}}$. $\mathbf{PM}$, the output of DSOM, is the grid approximation of the probability distribution of the pose. \rj{We discretize the state space $(x,y,\theta)$ by $[H_{env},W_{env},K]$. Assuming the environment is $x$ meters in length and $y$ meters in width, the entry $\left(i,j,k\right)$ in $\mathbf{PM}\in \mathbb{R}^{H_{env}\times W_{env}\times K}$ stands for the probability that the robot gets the observation $\mathbf{M}_{scan}$ on the exact pose $\left(i\times\frac{x}{H_{env}},j\times\frac{y}{W_{env}},k\times\frac{2\pi}{K}\right)$.} DSOM consists of four main parts: (1) Observation Encoder (2) Environment Encoder (3) Similarity Computation (4) Distribution Decoder. \hyperref[sec: Architecture]{Section A} shows details in these four parts and the loss function. We show the ground truth generation in \hyperref[sec: Ground Truth Generation]{Section B} and in \hyperref[sec: Multi-Modal Effect]{Section C}, we explain why DSOM can approximate the multimodal probability distribution.


\subsection{Architecture}
\label{sec: Architecture}
\noindent\textbf{Observation Encoder. }As described in Equation \eqref{equation for observation feature extraction}, $F_{enc}^{scan}$ is first deployed to encode $\mathbf{M}_{scan}$ to a feature map and then a Multi-Layer Perceptron ($MLP$) transforms this feature map to the feature representation $\mathbf{S}\in\mathbb{R}^{K\times D}$. We can see that there are $K$ feature vectors in $\mathbf{S}$, each of which is related to the feature at each discretized rotation angle. As illustrated in Figure \ref{fig:DMOM}, $F_{enc}^{scan}$ consists of several encoding blocks, each of which contains convolutional layers, batch normalization layers, non-linear activation layers and pooling layers.
\begin{equation}
\mathbf{S}=MLP\left(F_{enc}^{scan}\left(\mathbf{M}_{scan}\right)\right)
\label{equation for observation feature extraction}
\end{equation}

\noindent\textbf{Environment Encoder. }As Equation \eqref{equation for map feature extraction} shows, $F_{enc}^{env}$, which is similar to $F_{enc}^{scan}$, transforms the map of the environment $\mathbf{M}_{env}\in \mathbb{R}^{H_{env}\times W_{env}}$ to its feature map $\mathbf{M}\in \mathbb{R}^{\frac{H_{env}}{8}\times \frac{W_{env}}{8}\times D}$. 
\begin{equation}
\mathbf{M}=F_{enc}^{env}\left(\mathbf{M}_{env}\right)
\label{equation for map feature extraction}
\end{equation}

\noindent\textbf{Similarity Computation. }Cosine similarity is deployed to compute the similarity between $\mathbf{S}$ and $\mathbf{M}$, resulting in the similarity feature map $\mathbf{SIM}\in \mathbb{R}^{\frac{H_{env}}{8}\times \frac{W_{env}}{8}\times K}$. As described in Equation \eqref{equation for similarity computation}, $\mathbf{S}_k\in \mathbb{R}^D$ is the $k^{th}$ feature vector in $\mathbf{S}$, $\mathbf{M}_{i,j}\in \mathbb{R}^D$ is the feature vector on pixel $(i,j)$ on $\mathbf{M}$ and $\epsilon$ is a very small number to avoid being divided by zero. $\mathbf{SIM}_{i,j,k}$ indicates the $(i,j,k)$ entry in $\mathbf{SIM}$.
\begin{equation}
\mathbf{SIM}_{i,j,k}=\frac{{\mathbf{S}_k}^T\  \mathbf{M}_{i,j}}{max\left(\|\mathbf{S}_k\|_2\cdot\|\mathbf{M}_{i,j}\|_2,\epsilon\right)}
\label{equation for similarity computation}
\end{equation}

\noindent\textbf{Distribution Decoder. }In Equation \eqref{equation for pmap generation}, $F_{dec}$ and a softmax activation $\sigma$ lift $\textbf{SIM}$ to $\textbf{PM}$, which stands for the grid approximation of probability distribution over the state space. $F_{dec}$ consists of several decoding blocks, each of which contains convolutional layers, batch normalization layers, non-linear activation layers and an unpooling layer. The unpooling indexes are the same as the pooling indexes in the environment encoder.
\begin{equation}
\mathbf{PM}=\sigma\left(F_{dec}\left(\mathbf{SIM}\right)\right)
\label{equation for pmap generation}
\end{equation}

\noindent\textbf{Loss Function. }Kullback-Leibler divergence Loss \cite{van2014renyi} (KLD Loss) is deployed to guide the training process. A unimodal probability distribution at the exact pose where the robot obtains the observation is computed as the ground truth, $\mathbf{GT}$. Equation \eqref{equation for loss} illustrates how KLD Loss works, where $\mathbf{GT}_{i,j,k}$ and $\mathbf{PM}_{i,j,k}$ respectively indicate the $(i,j,k)$ entry in $\mathbf{GT}$ and $\mathbf{PM}$.
\begin{equation}
Loss=\sum_{i=1}^{H_{env}}\sum_{j=1}^{W_{env}}\sum_{k=1}^{K}\left[\mathbf{GT}_{i,j,k} \log{\frac{\mathbf{GT}_{i,j,k}}{\mathbf{PM}_{i,j,k}}}\right]
\label{equation for loss}
\end{equation}

\subsection{Ground Truth Generation}
\label{sec: Ground Truth Generation}
To generate $\mathbf{GT}\in\mathbb{R}^{H_{env}\times W_{env}\times K}$ for a given pose $\left(x,y,\theta\right)$, we begin with an all-zero matrix for $\mathbf{GT}$. Then a GaussianBlur is deployed on a one-hot ground truth matrix of $\left(x,y\right)$, resulting in $\mathbf{GT}'\in \mathbb{R}^{H_{env}\times W_{env}}$. After that, we find the nearest two orientation values ($\theta_1$ and $\theta_2$) to $\theta$ in the discretized angle space. Then a linear interpolation is deployed on these two values as below, where $I_{\theta_{1}}$ indicates the discretized index for $\theta_{1}$ and $\mathbf{GT}_{:,:,I_{\theta_{1}}}$ are all the entries of which third index is $I_{\theta_{1}}$
\begin{equation}
\begin{split}
\mathbf{GT}_{:,:,I_{\theta_{1}}}&=|\frac{\theta_1-\theta}{\theta_1-\theta_2}|\\[1mm]
\mathbf{GT}_{:,:,I_{\theta_{2}}}&=1 - \mathbf{GT}_{:,:,I_{\theta_1}}\\[1mm]
\end{split}
\label{GT generation: linear}
\end{equation}
We aggregate both the position and orientation ground truth as described in Equation \eqref{GT generation: aggregation} and then a normalization is deployed on $\mathbf{GT}$ leading to the final ground truth. $\mathbf{GT}_{i,j,k}$ is the value on the entry (i,j,k) of $\mathbf{GT}$.
\begin{equation}
\mathbf{GT}_{i,j,k}=\mathbf{GT}_{i,j,k}\cdot\mathbf{GT}'_{i,j}
\label{GT generation: aggregation}
\end{equation}

\subsection{Multimodal Effect in DSOM}
\label{sec: Multi-Modal Effect}
While there might exist a set of similar observations $\{\mathbf{M}_{scan}^{l}|l=1,2,...,m\}$ obtained at a set of different states $\{s^{l}|l=1,2,...,m\}$ in the training set, the ground truth $PM$ is only unimodal distribution. Why can the unimodal ground truth guide the network to generate a multimodal probability distribution? \rj{We here prove the case in which $\{\mathbf{GT}^l|l=1,2,...,m\}$ are one-hot distribution, as described in Equation \eqref{eq: GT} where $\mathbf{GT}_{i,j,k}^{l}$ is the $\{i,j,k\}$ entry in $\mathbf{GT}^l$, $ i_{g t}^l,\ j_{g t}^l,\ k_{g t}^l$ are respectively the ground truth entry for position and orientation. We further assume that all the $\mathbf{M}_{scan}^{l}$ are the same for simplicity. The conclusion is that the sum of the KLD Losses of these $m$ samples is minimized only when the output $\mathbf{PM}$ is an even-distributed multimodal distribution on these $m$ poses.
	
	\begin{equation}
	\begin{array}{l}
	\Delta(i, j)=\left\{\begin{array}{l}
	1, \text { if } i=j \\
	0, \text { otherwise }
	\end{array}\right. \\
	\mathbf{GT}_{i,j,k}^{l}=\Delta\left(i, i_{g t}^l\right) \cdot \Delta\left(j, j_{g t}^l\right) \cdot \Delta\left(k, k_{g t}^l\right)
	\end{array}
	\label{eq: GT}
	\end{equation}
}

\noindent \textit{Proof. }As $\mathbf{M}_{scan}^{l}$ are the same, all the $\mathbf{S}$ generated by observation encoder are identical while environment encoder obtains the same feature map $\mathbf{M}$. Thus all the $\mathbf{PM}^l$ are the same, denoted as $\mathbf{PM}$. The sum of losses is described in Equation \eqref{Sum of loss}, where $\mathbf{PM}_{s^l}$ indicates value on the entry $s^l$ of $\mathbf{PM}$. 
\begin{equation}
\begin{split}
Loss&=\sum_{l=1}^{m}\left\{\sum_{i=1}^{H_{env}}\sum_{j=1}^{W_{env}}\sum_{k=1}^{K}\left[\mathbf{GT}_{i,j,k}^l \log{\frac{\mathbf{GT}_{i,j,k}}{\mathbf{PM}_{i,j,k}}}\right]\right\}\\[1mm]
&=\sum_{l=1}^{m}-\log{\mathbf{PM}_{s^l}}
\end{split}
\label{Sum of loss}
\end{equation}

Thus the problem turns to:
\begin{equation}
\begin{aligned}
& \underset{\mathbf{\mathbf{PM}}}{\text{minimize}}
& & \sum_{l=1}^{m}-\log{\mathbf{PM}_{s^l}} \\
& \text{subject to}
& & \sum_{s}\mathbf{PM}_s=1
\end{aligned}
\end{equation}

After applying Lagrange multiplier to it, the problem turns to minimize the term below:
\begin{equation}
\underset{\mathbf{\mathbf{PM},\lambda}}{\text{minimize}}\ \ \ \ L=\sum_{l=1}^{m}-\log{\mathbf{PM}_{s^l}}+\lambda(1-\sum_{s}\mathbf{PM}_{s})
\end{equation}

Let the derivative of $L$ with respect to every entry in $\mathbf{PM}$ and $\lambda$ be 0, we get:
\begin{equation}
\begin{split}
\frac{1}{\lambda}&=\mathbf{PM}_{s^1}=\mathbf{PM}_{s^2}=...=\mathbf{PM}_{s^m}\\[1mm] 
\mathbf{PM}_s&=0,\ \ \forall s\notin \{s^{l}|l=1,2,...,m\}
\end{split}
\end{equation}

Also with $\sum\limits_{s}\mathbf{PM}_s=1$, we can get that only when Equation \eqref{conclusion} is satisfied, can the sum of these $m$ KLD Losses get minimized. Proof complete. $\blacksquare$ 

\begin{equation}
\mathbf{PM}_{s^1}=\mathbf{PM}_{s^2}=...=\mathbf{PM}_{s^m}=\frac{1}{m}
\label{conclusion}
\end{equation}

As training samples can cover part of the state space, it is obvious that the KLD Loss can guide the network to approximate the multimodal distribution.

\section{Adaptive Mixture MCL}
\label{sec: Adaptive Mixture MCL}

\rj{As the training samples cannot cover all the possible cases, $\mathbf{PM}$ sometimes fails to approximate the accurate probability distribution. The combination of traditional MCL and Dual MCL with DSOM provides a solution. However, the combination in \cite{thrun2001robust}, which deploys either MCL with probability $p$ or Dual MCL with probability $1-p$ in one iteration, is too simple. If the proposal distribution fails to capture the correct poses at some iteration and the algorithm happens to deploy Dual MCL, the algorithm will lose track of the robot pose. To compensate for this problem, we propose a novel filter, Adaptive Mixture MCL (AdaM MCL).}

In AdaM MCL, each particle has three elements $\left(s^i,w^i_{norm},w^i\right)$, where the first one indicates the pose of this particle, $w^i$ is the weight before normalization and $w^i_{norm}$ is the normalized weight. In every updating iteration, we first evaluate the degree we trust a particle by computing $T=\frac{w^i}{w_{*}}$, where $w_{*}$ is the output of the weighting model when given zero-difference in every direction as input. \rj{Please refer to Section A in Appendix for the detailed equation for $w_{*}$.}  According to $T$, we divide the whole set into $\mathcal{H}$ and $\mathcal{L}$. As there are obstacles around the robot, $w_i$ is never equal to $w_{*}$ even when the particle is at the same pose as that of the robot. To compensate for this situation, we introduce a threshold parameter $T_{cut}\in\left(0,1\right)$ and if $T>T_{cut}$, we directly add this particle to $\mathcal{H}$. If not, this particle is added to $\mathcal{H}$ with a probability of $T$ or $\mathcal{L}$ with a probability of $1-T$. Finally, we deploy traditional MCL update for $\mathcal{H}$ and Dual MCL update with $\mathbf{PM}$ for $\mathcal{L}$. \rj{Algorithm 1 in Section D in the Appendix shows how AdaM MCL works.}

\vspace{-0.1cm}

\section{Experiments and Results}
\label{sec: Experiments and Results}
In this section, we conduct experiments in synthetic and real scenes to investigate the ability of DSOM to approximate the multimodal distribution and generalize across scenes. Also, we compare AdaM MCL with previous methods on global localization and kidnapping problems.
\subsection{Experiments Setup}
\label{sec: Experiments Setup}
\noindent\textbf{Datasets. }We conduct experiments in both synthetic and real environments.

\begin{figure*}[h]
	\centering
	\subfigure[Scene 1]{
		\begin{minipage}[t]{0.16\linewidth}
			\centering
			\includegraphics[width=0.96in]{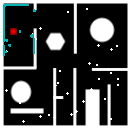}
		\end{minipage}%
	}%
	\centering
	\subfigure[Prediction 1]{
		\begin{minipage}[t]{0.16\linewidth}
			\centering
			\includegraphics[width=0.96in]{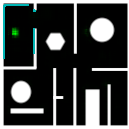}
		\end{minipage}%
	}%
	\centering
	\subfigure[Scene 2]{
		\begin{minipage}[t]{0.16\linewidth}
			\centering
			\includegraphics[width=0.96in]{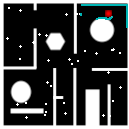}
		\end{minipage}%
	}%
	\centering
	\subfigure[Prediction 2]{
		\begin{minipage}[t]{0.16\linewidth}
			\centering
			\includegraphics[width=0.96in]{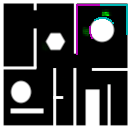}
		\end{minipage}%
	}%
	\centering
	\subfigure[Scene 3]{
		\begin{minipage}[t]{0.16\linewidth}
			\centering
			\includegraphics[width=0.96in]{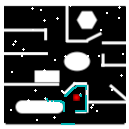}
		\end{minipage}%
	}%
	\centering
	\subfigure[Prediction 3]{
		\begin{minipage}[t]{0.16\linewidth}
			\centering
			\includegraphics[width=0.96in]{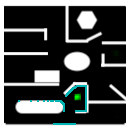}
		\end{minipage}%
	}%
	\vspace{-0.2cm}
	\caption{Visualization results on the output of DSOM. As shown in (a)(c)(e), observation with obstacles around is indicated by blue lines. Red squares in (a)(c)(e) indicate the ground truth position. The input of DSOM are observation in (a)(c)(e) and a map without obstacles. Green clusters in (b)(d)(f) show the predicted probability peaks. Blue and purple lines in (b)(d)(f) show the observation on the predicted probability peaks.}
	\label{fig: results on Multi-modal}
\end{figure*}

\begin{figure}[t]
	\centering
	\includegraphics[height=0.85\linewidth]{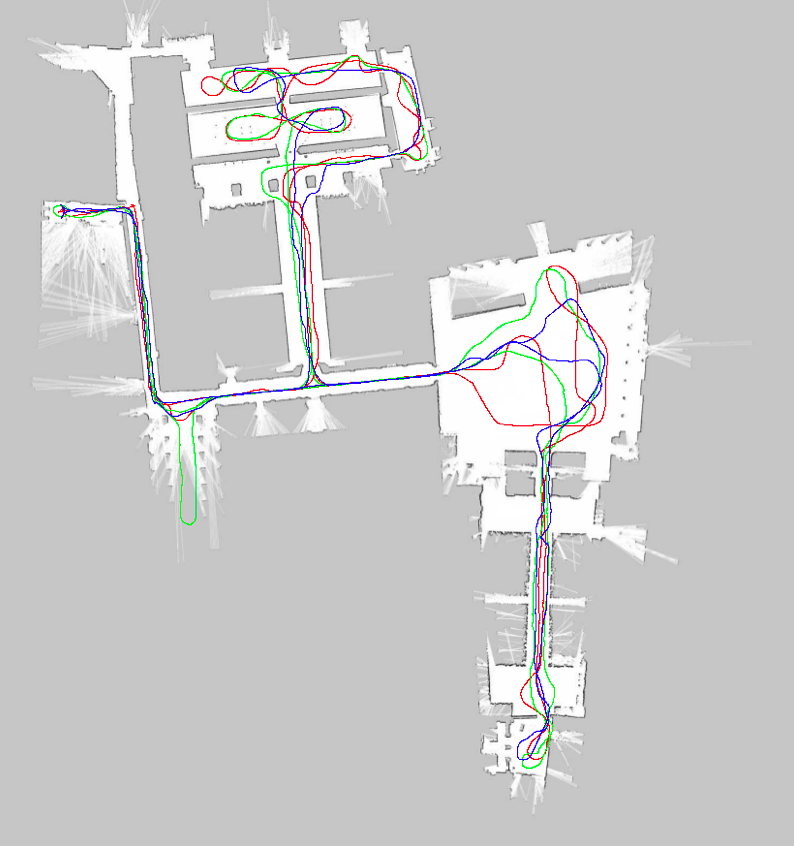}
	\caption{The map of UPO dataset \cite{ramon2014navigating} and multiple trajectories in this dataset. We \rj{randomly} select three testing clips among them.}
	\label{fig: results on UPO}
	\vspace{0.5cm}
\end{figure}

\begin{itemize}[leftmargin=*]
	\item \textit{Synthetic Environments. }We manually create 20 maps \rj{(43 meters in both length and width)}. We collect training data in 18 of them and testing sequences in the other 2 maps. To train DSOM, we add random obstacles in the environments and simulate laser to get the observation. What we feed into DSOM is the original map without obstacles and the observation. For the testing sequential data, we put unknown obstacles or remove part of the environment along the path.
	\item \textit{Real Environments. }We conduct experiments on two benchmark real-world datasets: the Royal Alcazar of Seville dataset (UPO) \rj{(150 meters in length and 100 meters in width)} \cite{ramon2014navigating} and the Rawseeds indoor dataset collected in the University di Milano-Bicocca (Bicocca) \cite{ceriani2009rawseeds,bonarini2006rawseeds}. For the Bicocca dataset, we divide the whole scene into two smaller parts \rj{(40 meters in length and 30 meters in width)} for experiments. To train DSOM, we add random obstacles to the map given by these datasets (SLAM-based or floorplan) and collect training data using simulating laser. With these synthetic data, we train DSOM, after which it is applied to real observation in the same scene. A map of the environment where the UPO dataset is collected is shown in Figure \ref{fig: results on UPO}.
\end{itemize}

\noindent\textbf{Baselines. }There are four baselines in our experiments.

\begin{itemize}[leftmargin=*]
	\item \textit{MCL} with random sampling \cite{thrun2001robust}.
	\item \textit{Mixture MCL} \cite{thrun2001robust} whose sampling model is trained by data collected in the testing scene with random obstacles.
	\item \textit{Mixture MCL*} whose sampling model is trained by data containing the testing sequences except for the exact pose at a given timestamp.
	\item \textit{PFRNN }\cite{ma2019particle}\footnote{We use the implementation from the github repository of the author: https://github.com/Yusufma03/pfrnns} the most recent and effective deep-learning-based particle filter.
	\rj{\item  \textit{PF-Net} \cite{karkus2018particle}: the particle-filter network.}
\end{itemize}

\noindent For \textit{MCL} and \textit{Mixture MCL}, we follow the implementation in \cite{thrun2002probabilistic,quigley2009ros} and also try our best to tune the parameters in these two baselines for better performance than the original parameter set in \cite{thrun2002probabilistic,quigley2009ros}. For \textit{Mixture MCL*}, we note here that the sampling model is ``cheating'' because an algorithm should not have any prior about the testing sequences. \rj{PFRNN and PF-Net are trained by data sequences collected in the same environment where they are tested.} The random sampling rate of traditional MCL is set to $0.2$. Particle number is set to 200 for all algorithms in synthetic datasets and 500 for all real datasets. 

\vspace{0.1cm}

\noindent\textbf{DSOM and AdaM MCL. } The network structure in DSOM, as shown in \hyperref[sec: DSOM detailed settings]{\rj{Section B in Appendix}}, remains the same in different experiments (different environment sizes) to demonstrate its scalability. $T_{cut}$ is set to $0.6$ for all synthetic environments and the UPO dataset. For the Bicocca dataset, $T_{cut}$ is set to $0.5$ because the environment is more dynamic than others.

\vspace{0.1cm}
\rj{\noindent\textbf{About the GPU Memory Consumption and Time Consumption: }As the maps of the environments grow, the GPU memory consumption becomes larger. To make a fair comparison of the three deep-learning-based methods, we limit the available GPU to only one TITAN GPU with 12 Gigabytes memory. By scaling the resolution of the map, we can train and evaluate the models with limited resources. And typically, inference for one iteration in DSOM requires only 0.02 seconds while PFRNN and PF-Net require only 0.08 seconds, all of which are negligible. Thus, we will use the number of updating iterations before convergence to compare the converging speed of all the methods.}

\vspace{0.1cm}
\noindent\textbf{Evaluation Metric. }For each testing sequence and method, we run experiments for 100 times. Then the error of the estimated position $\left(x_{est},y_{est},\theta_{est}\right)$ is computed at each updating iteration: E$_{pos}=\sqrt{\left(x_{est}-x_{gt}\right)^2+\left(y_{est}-y_{gt}\right)^2}$ and E$_{\theta}=\sqrt{\left(\theta_{est}-\theta_{gt}\right)^2}$. Thus at each updating iteration on each testing sequence, we have 100 error values for each method. To show the converging process of each method, we compute the mean value as well as 95\% confidence interval of the 100 error values E$_{pos}$ at each updating step and plot the evolution of estimation error of all methods on the same chart to compare their performance. Because each update iteration in all the methods we use can be finished in real-time, we care more about how long the robot travels before the algorithm converges than the time needed to converge. To evaluate how fast and stable an algorithm converges, we first set a condition for converging to the accurate pose: If E$_{pos}$ is lower than 2 meters and the error in orientation E$_{\theta}$ is lower than 10 degrees for 5 consecutive updating steps, we consider that the algorithm converges to the accurate pose. If an algorithm converges, $STEPS$ is computed as the number of steps the algorithm requires to converge. 



\subsection{Model Validation}
\noindent\textbf{Case Study of the multimodal effect in DSOM.}
In Figure \ref{fig: results on Multi-modal}, it can be found that DSOM is able to take the observation and map without obstacles as inputs and generate various numbers of probability peaks, including the ground truth, which demonstrates DSOM as a good sampling model. A visualization of the output of DSOM on Bicocca dataset can be found in \hyperref[Visualization of the output of DSOM on Bicocca dataset.]{Appendix C}.

\begin{figure}[H]
	\begin{center}
		\includegraphics[width=0.9\linewidth]{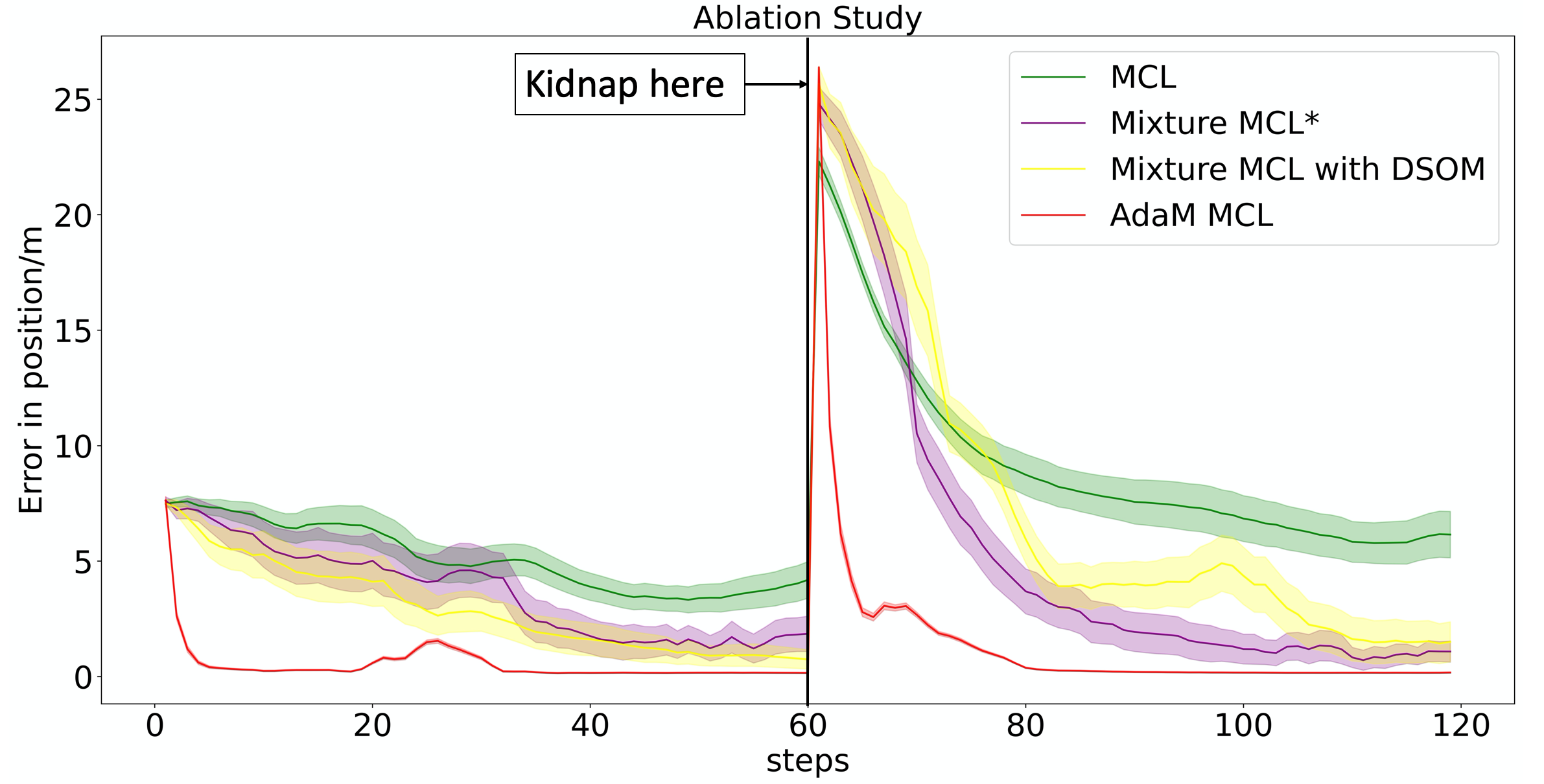} %
	\end{center}
	\vspace{-0.2cm}
	\caption{Results on the ablation study experiment.}
	\label{fig: results on Ablation Study}
	\vspace{-1mm}
\end{figure}

\vspace{0.1cm}
\noindent\textbf{Ablation Study.}
We compare four methods in a testing synthetic environment: MCL, Mixture MCL*, Mixture MCL with DSOM as the sampling model and AdaM MCL. The results for global localization and kidnapping with these four methods are shown in Figure \ref{fig: results on Ablation Study}. It can be found that Mixture
\begin{table}[H]
	\centering
	\vspace{-0.05cm}
	\caption{Parameter sensitivity experiment on $T_{cut}$}
	\label{Table: Parameter sensitivity experiment on T cut}
	\begin{tabular}{ccccc}
		\hline
		\multirow{2}{*}{$T_{cut}$} & \multicolumn{2}{c}{Global Localization} & \multicolumn{2}{c}{Kidnapping} \\
		& $E_{pos}/m$         & $E_{\theta}/rad$        & $E_{pos} /m $  & $E_{\theta}/rad$    \\ \hline
		0.1                          & 0.299             &        0.017             & 0.188        &         0.031        \\
		0.3                          & 0.190             &         0.019            & 0.150        &         0.021        \\
		0.5                          & 0.163             &            0.018         & 0.169        &          0.018       \\
		0.7                          & 0.155             &         0.015            & 0.286        &            0.053     \\
		0.9                          & 0.725             &             0.023        & 1.133        &          0.547       \\ \hline
	\end{tabular}
	\vspace{-5mm}
\end{table}
\begin{table}[H]
	\vspace{0.3cm}
	\centering
	\caption{Parameter sensitivity experiment on particle number.}
	\label{Table: Parameter sensitivity experiment on particle number}
	\begin{tabular}{ccccc}
		\hline
		\multirow{2}{*}{Particle Number} & \multicolumn{2}{c}{Global Localization} & \multicolumn{2}{c}{Kidnapping} \\
		& $E_{pos}/m$         & $E_{\theta}/rad$        & $E_{pos} /m $  & $E_{\theta}/rad$    \\ \hline
		200                          & 0.153             &             0.022        & 0.128        &         0.022        \\
		500                          & 0.088             &           0.017          & 0.089        &            0.016     \\
		1000                          & 0.060             &              0.016       & 0.072        &          0.007       \\ \hline
	\end{tabular}
	\vspace{-2mm}
\end{table}
\noindent MCL with DSOM achieves slightly better performance in the global localization problem while Mixture MCL* outperforms a bit in the kidnapping problem. The only difference between them lies in the sampling model and as discussed previously, the sampling model in Mixture MCL* is trained by testing data while DSOM is trained by data collected in other environments. This demonstrates that DSOM can generalize across synthetic environments and provide an accurate probability distribution over the state space for sampling (similar performance to a ``cheating'' sampling model). Also, it is found that AdaM MCL converges faster and achieves a more accurate estimation than Mixture MCL with DSOM, demonstrating the effectiveness of AdaM MCL.

\vspace{0.1cm}

\noindent\textbf{Parameter Sensitivity Experiment.}
\label{sec: Parameter Sensitivity Experiment}
We conduct parameter sensitivity experiments on two important parameters in AdaM MCL: $T_{cut}$ and the number of particles while other parameters are fixed. Results are shown in Table \ref{Table: Parameter sensitivity experiment on T cut} (the number of particles is set to 200) and \ref{Table: Parameter sensitivity experiment on particle number} ($T_{cut}$ is set to 0.6). It can be found that the estimation accuracy is relatively stable when the two parameters vary, which demonstrates that AdaM MCL is not sensitive to parameter settings.



\subsection{Comparisons to Baselines in Global Localization and Kidnapping}
\label{sec: Results on Global Localization and Kidnapping}
\begin{figure}[t]
	\centering
	\subfigure[Global Localization.]{
		\begin{minipage}[t]{0.9\linewidth}
			\centering
			\includegraphics[height=1.2in]{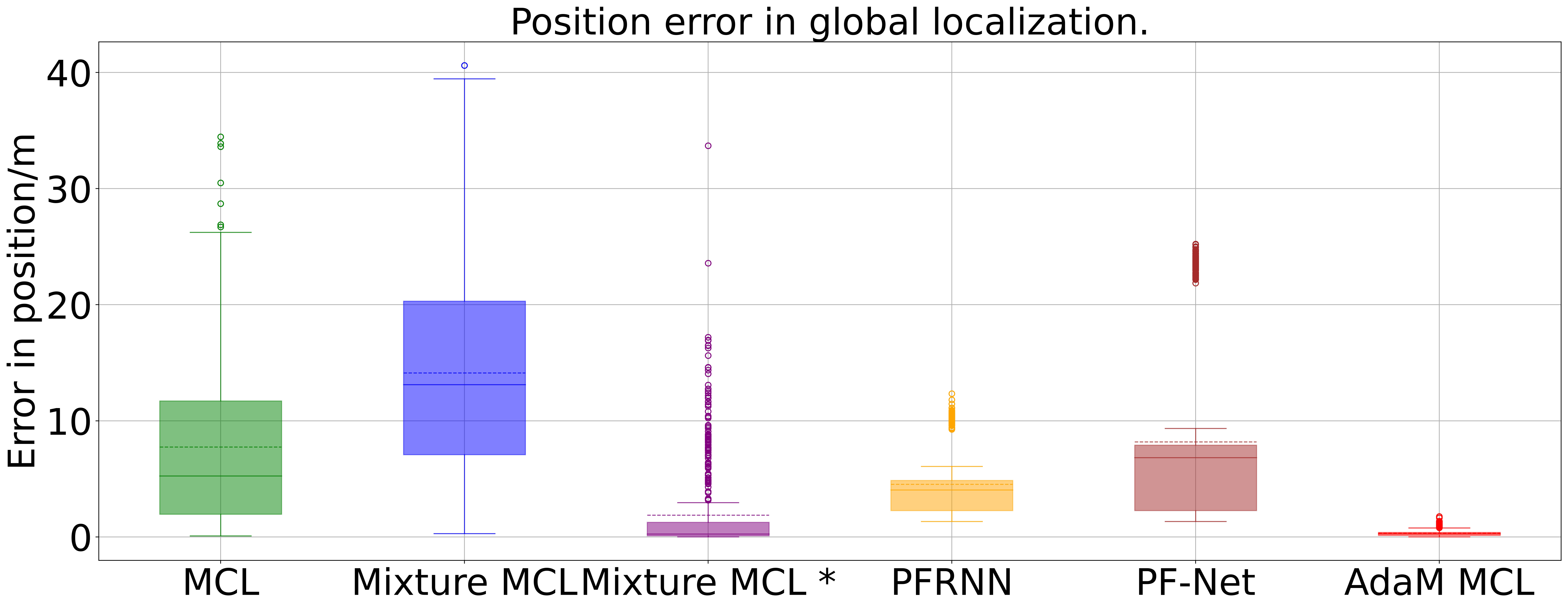}
		\end{minipage}%
	}%
	\\
	\centering
	\subfigure[\rj{Kidnapping}]{
		\begin{minipage}[t]{0.9\linewidth}
			\centering
			\includegraphics[height=1.2in]{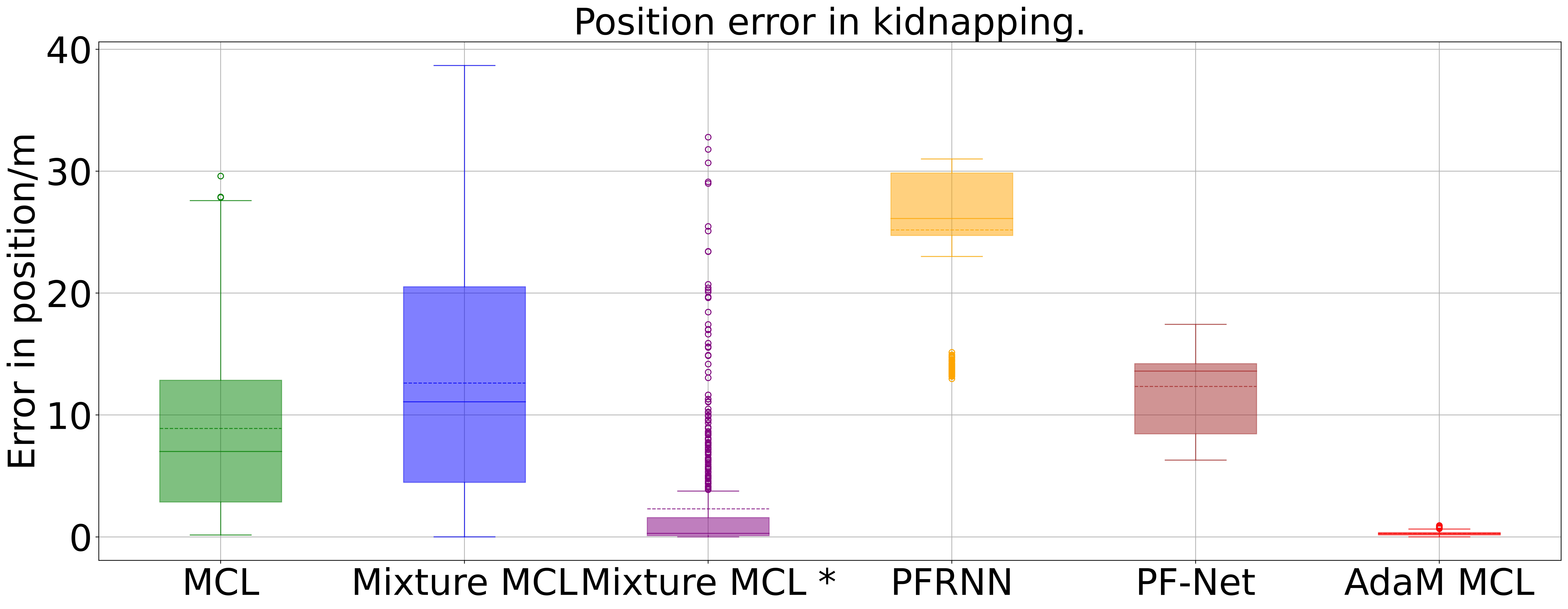}
		\end{minipage}%
	}%
	\vspace{-0.2cm}
	\caption{\rj{Results of E$_{pos}$ in synthetic environments.}}
	\label{results: E_pos}
\end{figure}

\begin{figure}[t]
	\centering
	\subfigure{
		\begin{minipage}[t]{0.96\linewidth}
			\centering
			\includegraphics[width=3in]{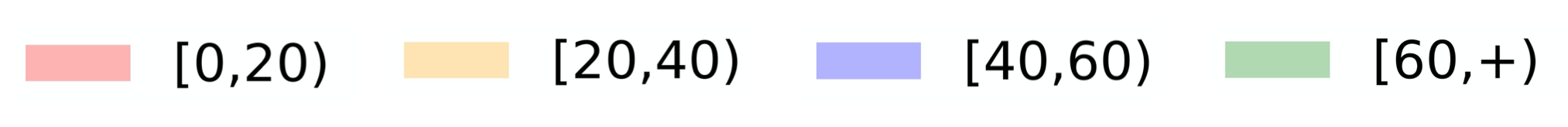}
		\end{minipage}%
	}
	\vspace{-0.3cm}
	\\
	\centering
	\subfigure[\small{Global Localization.}]{
		\begin{minipage}[t]{0.48\linewidth}
			\centering
			\includegraphics[width=1.6in]{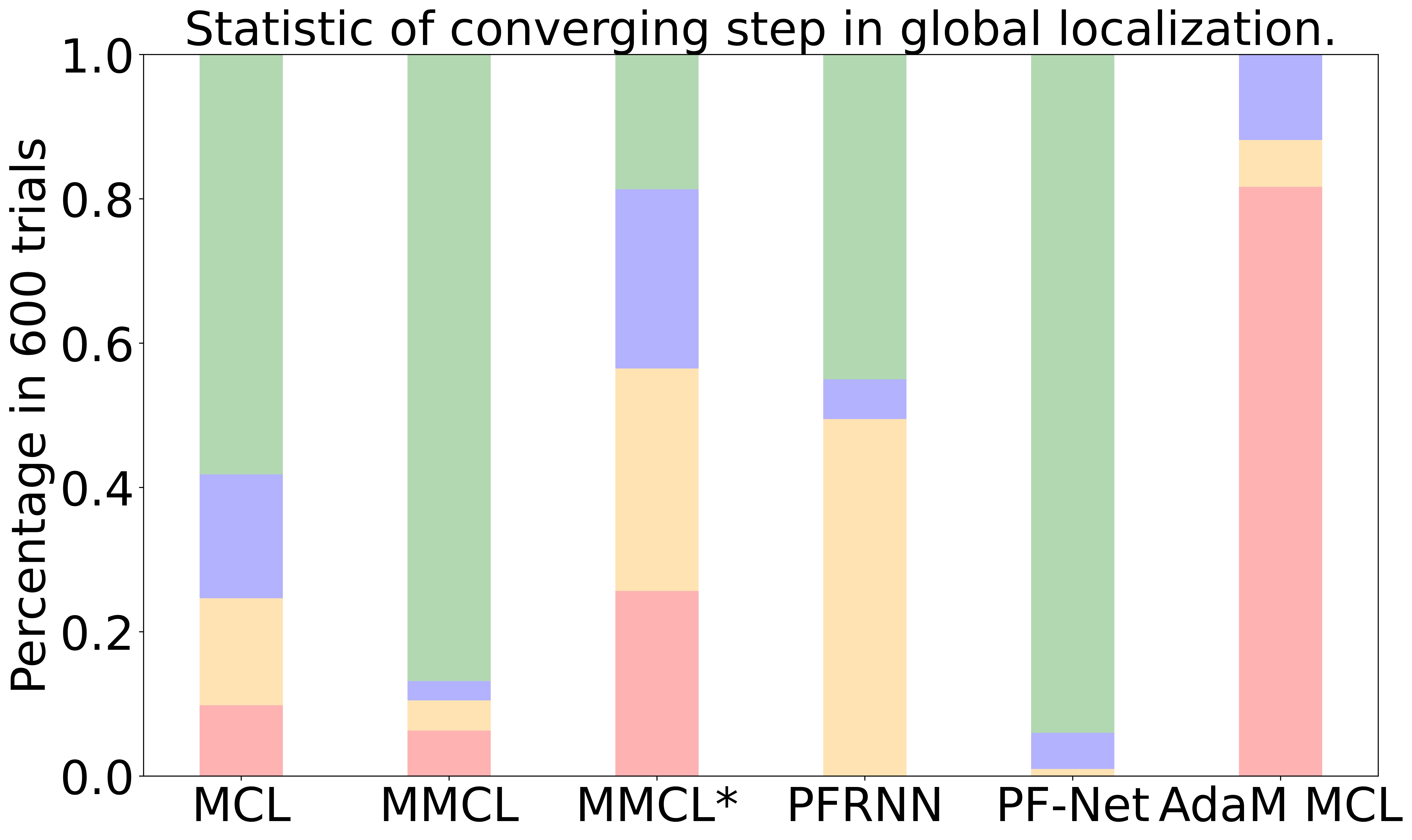}
		\end{minipage}%
	}%
	\subfigure[\small{Kidnapping.}]{
		\begin{minipage}[t]{0.48\linewidth}
			\centering
			\includegraphics[width=1.6in]{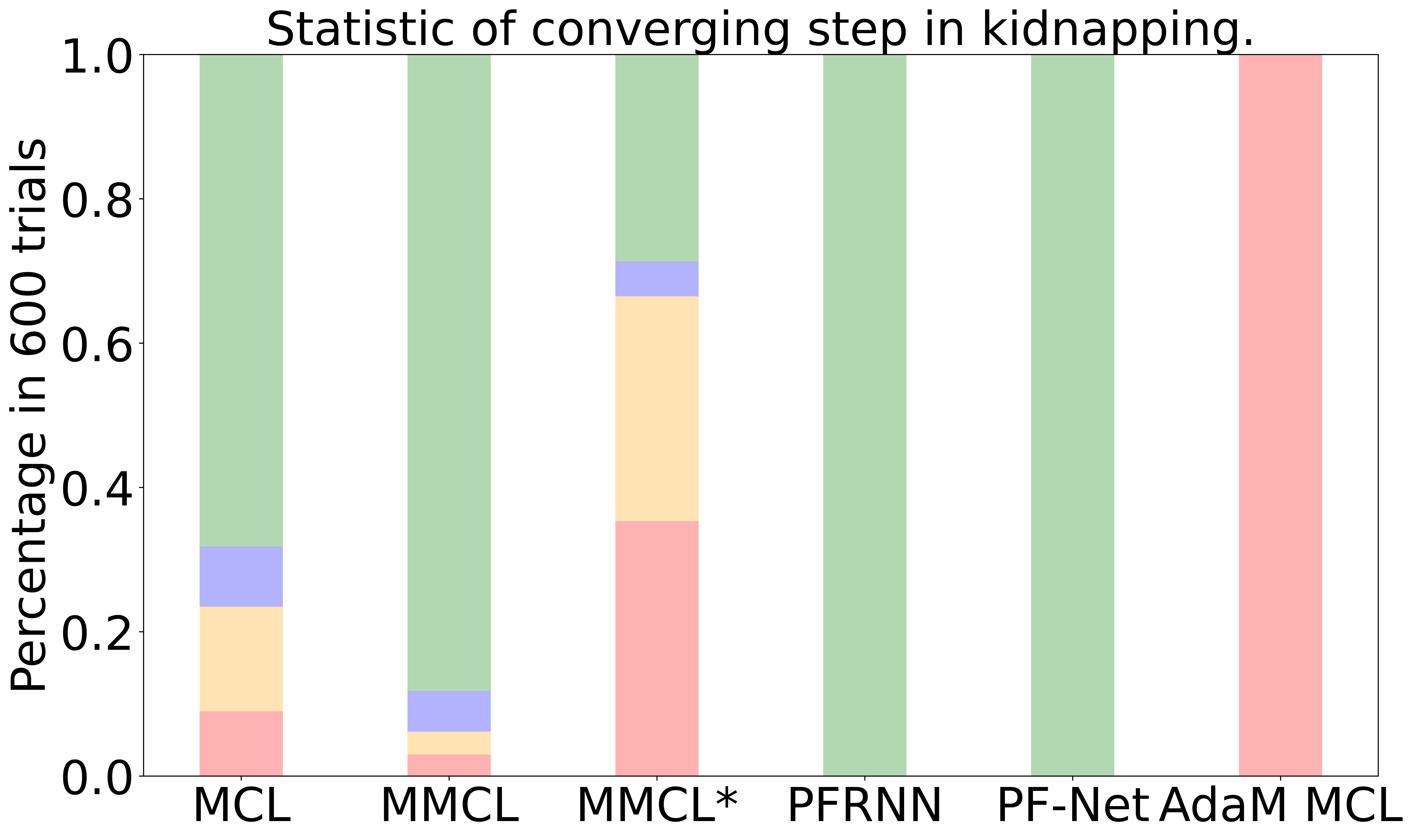}
		\end{minipage}%
	}%
	
	%
	\caption{\rj{Results for converging steps among 600 synthetic experiments. Red indicates the percentage that an algorithm converges in 0 to 20 steps among 600 experiments and it is similar for other colors. MMCL stands for Mixture MCL.}}
	\vspace{0.2cm}
	\label{fig: results on converging steps in synthetic environments.}
\end{figure}

\noindent\textbf{Train in some synthetic maps and test in ``unseen'' synthetic maps. }Six sequences of synthetic data are collected in two synthetic maps to evaluate the performance of the three methods, leading to 600 experiments for each method. For each sequence of testing data, we first start from global localization and then kidnap the robot for one time. Results for converging accuracy, i.e. E$_{pos}$, are  ``box-plotted'' in Figure \ref{results: E_pos}. For converging steps, we divide the statistic into 4 intervals and compute the percentage that each method converges within the respective interval. A normalized histogram about converging steps is shown in Figure \ref{fig: results on converging steps in synthetic environments.}.

It can be found that AdaM MCL achieves significantly better estimation results than all the baselines. Also, AdaM MCL achieves the highest converging rate and requires the least steps to converge. Note that DSOM is only trained by data collected in other environments while the sampling model in Mixture MCL* is trained by testing data and PFRNN as well as PF-Net is trained by data collected in the testing environments. These demonstrate that DSOM is able to generalize across different synthetic environments and AdaM MCL can accurately localize the robot in ``unseen'' synthetic scenes. 

\vspace{0.1cm}

\begin{figure*}[t]
	\centering
	\begin{minipage}[t]{0.96\linewidth}
		\centering
		\includegraphics[height=0.25in]{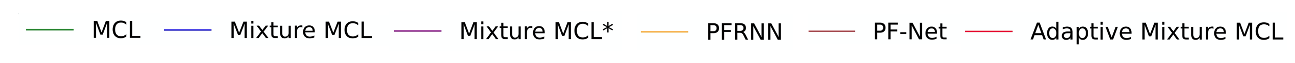}
	\end{minipage}%
	\vspace{-0.3cm}
	\centering
	\subfigure[Experiment on UPO dataset.]{
		\begin{minipage}[t]{0.31\linewidth}
			\centering
			\includegraphics[width=2.25in]{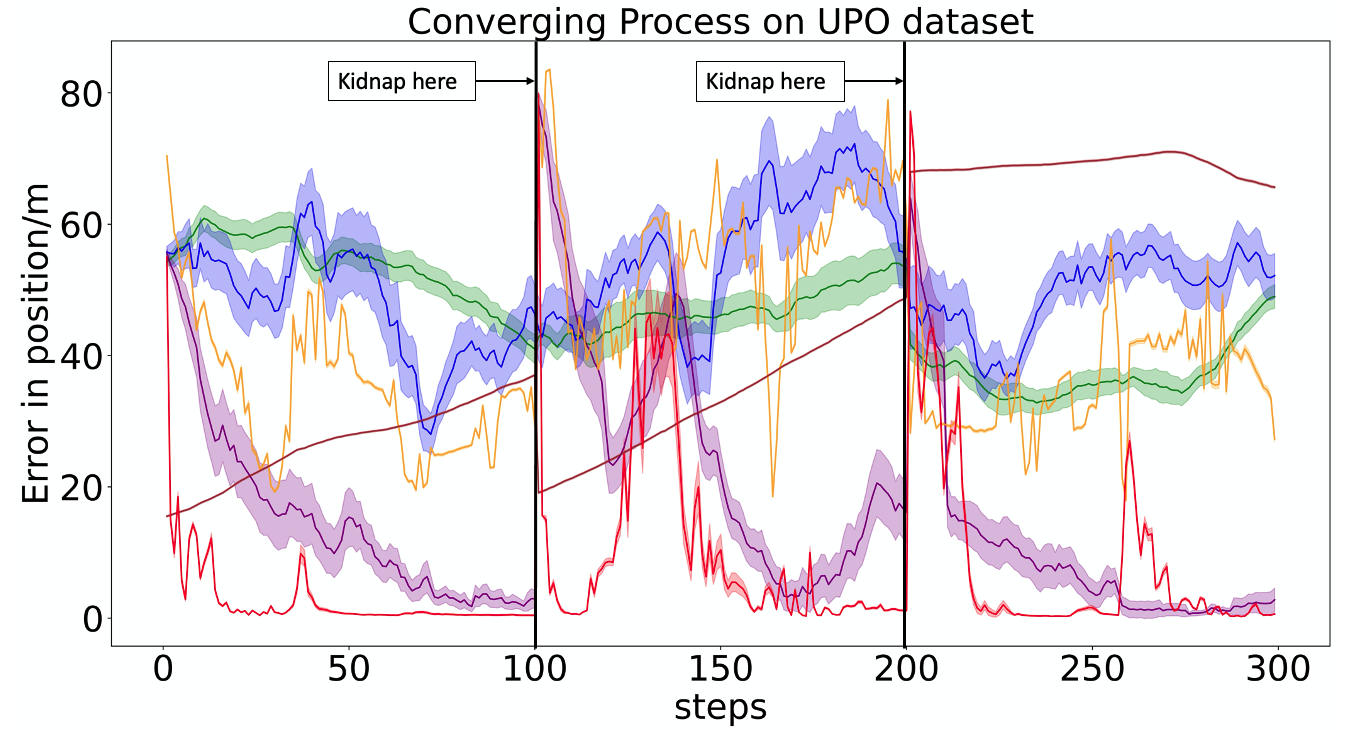}
		\end{minipage}%
		\label{key}	}%
	\centering
	\subfigure[Experiment on part 1 of Bicocca dataset.]{
		\begin{minipage}[t]{0.31\linewidth}
			\centering
			\includegraphics[width=2.25in]{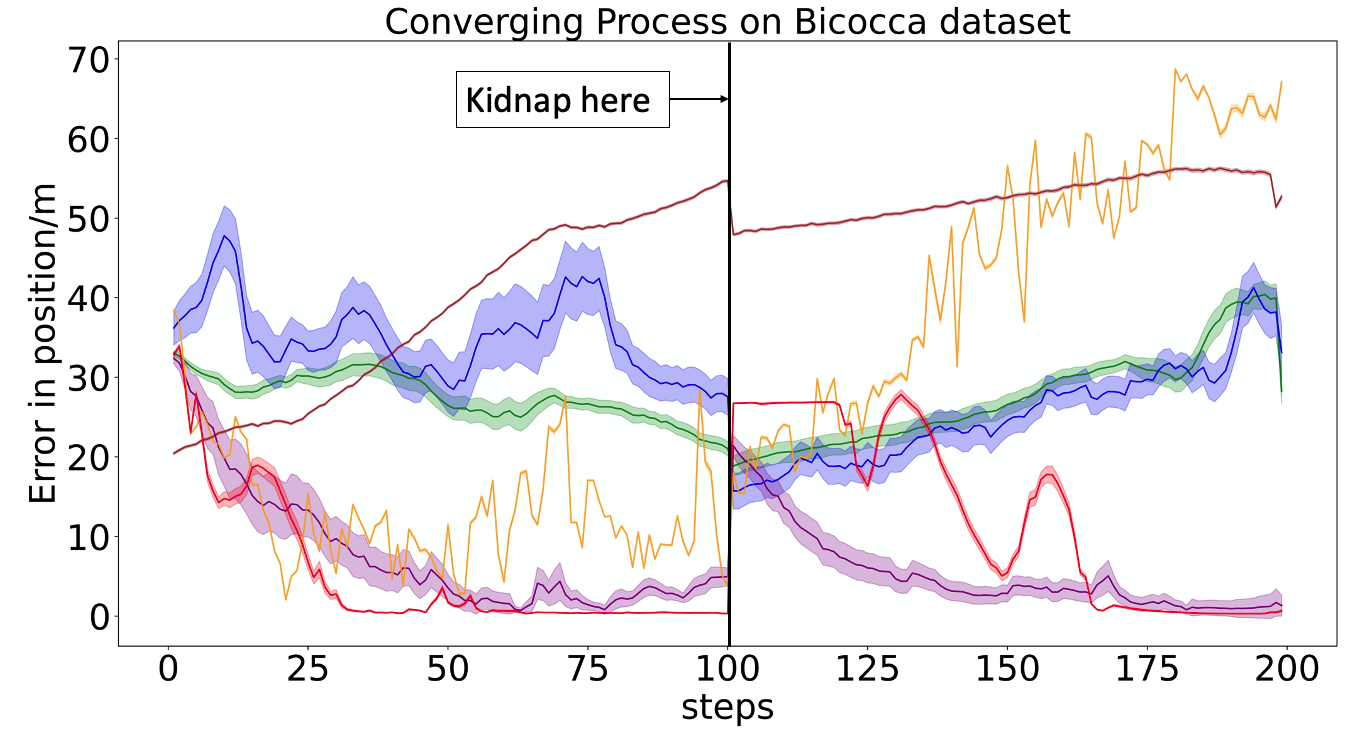}
		\end{minipage}%
	}%
	\centering
	\subfigure[Experiment on part 2 of Bicocca dataset.]{
		\begin{minipage}[t]{0.31\linewidth}
			\centering
			\includegraphics[width=2.25in]{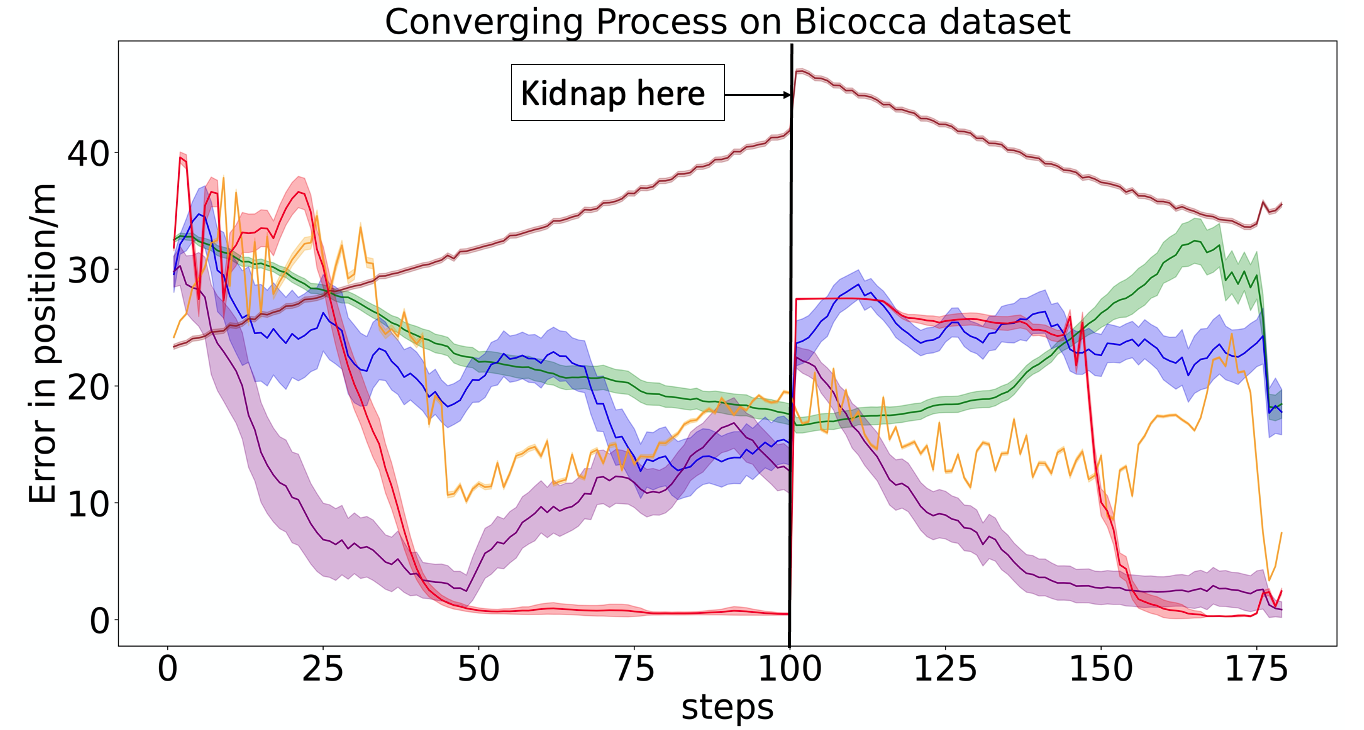}
		\end{minipage}%
	}%
	\caption{\rj{Results on real datasets.}}
	\label{fig: results on real datasets}
\end{figure*}

\noindent\textbf{Train with simulation observation on SLAM-based or floorplan maps and test on real observation. }As shown in Figure \ref{fig: results on real datasets}, it can be found that in real scenes, AdaM MCL with DSOM, which is trained by synthetic laser range data in the given blueprint or SLAM-based map, outperforms all the baseline on estimation accuracy, even Mixture MCL* whose sampling model is trained by testing data. These results indicate that DSOM can learn from synthetic laser data in the map of a real environment (SLAM-based or blueprint) and generalize to real observation, which helps AdaM MCL attain accurate estimation of the robot pose in real environments. This is very promising \rj{because} with a SLAM-based or blueprint map, which is always available, we can quickly generate simulation observations to train DSOM and then deploy AdaM MCL with DSOM to accurately localize the robot in the real scenes.

\section{Conclusion}
\label{sec: conclusion}
In this paper, we propose a samplable deep multimodal observation model, DSOM, as well as AdaM MCL, which is a novel localization filter for global localization and kidnapping problems. Our extensive experiments demonstrate that DSOM is able to generalize across different synthetic environments. Moreover, it can generalize to real environments when trained on only simulated observations from SLAM-based or even blueprint maps. \rj{Experiment results in both synthetic and real environments show that AdaM MCL equipped with DSOM achieves superior performance than previous methods on both synthetic and real datasets in global localization and kidnapping problems.}

\rj{In the future, there are two interesting directions that might stem from this work. The first is to transfer models trained in synthetic environments to real environments, whose difficulties stem from the considerable difference between synthetic and real environments. This can be achieved by several possible methods: retraining models on real datasets, which might accelerate the training process, or finding a suitable way to augment synthetic data to make it possible to directly transfer models trained on synthetic data to real data. The second direction is to extend the work to localization in 3D space with 3D observation information like point clouds or RGBD images. Recent deep learning techniques for processing 3D information such as \cite{qi2017pointnet++} can be incorporated into DSOM to extract robust features from both 3D maps and 3D observations. Then we can follow the same pipeline to decode a probability distribution for sampling poses.}


%

\appendices

\rj{
	\section{Transition and Observation Function}
	
	Given a particle $\left(s^i_{t-1},w^i_{t-1}\right)$, the motion $a_{t-1}$ and the observation $o_t$, the transition model is described in Equation \eqref{eq: transition model}, where $\sigma$ is the variance of motion infomation.
	
	\begin{equation}
	s^i_t = s^i_{t-1} + a_{t-1} + N(0,\sigma)
	\label{eq: transition model}
	\end{equation}
	
	Observation function is described in Equation \eqref{eq: observation model}, where $\mathbf{D}$ is the number of directions a 2D lidar perceives, $(L^{i}_{o}$ and $(L^{i}_{p}$ are the range in the $i^{th}$ direction of the real observation and the observation obtained at the pose of particle $p$ on the map. $\sigma$ is the variance of the lidar range finder and $L_{max}$ is the maximum range that the lidar can return. $\alpha$ and $\beta$ are parameters set as 0.9 and 0.1 respectively.
	
	\begin{equation}
	\text {Score}(L_o,L_p)=\sum_{i=1}^{\mathbf{D}}\left[\alpha \times \frac{e^{-\frac{\left(L^{i}_{o}-L^{i}_{p}\right)^{2}}{2 \sigma^{2}}}}{\sigma \sqrt{2 \pi}}+\beta \times \frac{1}{L_{\max }}\right]^{3}
	\label{eq: observation model}
	\end{equation}
	
	$w_{*}$ is the output of the weighting model when given zero-difference in every direction as inputs, that is $w_{*}=\text{Score}(L,L)$, where $L=L_o=L_p$}

\section{Detailed Settings in DSOM}
\label{sec: DSOM detailed settings}

\vspace{-3mm}
\begin{table}[H]
	\caption{Detailed Settings in DSOM}
	\centering
	\begin{minipage}[h]{0.48\linewidth}
		
		\begin{tabular}{cccc}
			\hline
			& Stride & Filters & Size  \\ \hline
			\multirow{10}{*}{\rotatebox{90}{Observation Encoder}} &  2      & 64      & (3,3) \\
			& 2      & 64      & (3,3) \\
			& 2      & 64     & (3,3) \\
			& 2      & 64     & (3,3) \\
			& 2      & 128     & (3,3) \\
			& 2      & 128     & (3,3) \\
			& 2      & 256     & (3,3) \\
			& 2      & 256     & (3,3) \\
			& 2      & 256    & (3,3) \\ 
			& 2      & 256    & (3,3) \\\hline
		\end{tabular}
	\end{minipage}
	\centering
	\begin{minipage}[h]{0.48\linewidth}
		\begin{tabular}{cccc}
			\hline
			& Stride & Filters & Size  \\ \hline
			\multirow{10}{*}{\rotatebox{90}{Distribution Decoder}} & 2      & 256     & (3,3) \\
			& 2      & 256     & (3,3) \\
			& 2      & 256     & (3,3) \\
			& 2      & 256     & (3,3) \\
			& 2      & 128     & (3,3) \\
			& 2      & 128     & (3,3) \\
			& 2      & 128     & (3,3) \\
			& 2      & 64      & (3,3) \\
			& 2      & 64      & (3,3) \\
			& 2      & 32      & (3,3) \\ \hline
		\end{tabular}
	\end{minipage}
	\\
	\begin{minipage}[t]{0.48\linewidth}
		\vspace{0.5cm}
		\begin{tabular}{cccc}
			\hline
			& Stride & Filters & Size  \\ \hline
			\multirow{9}{*}{\rotatebox{90}{Env. Encoder}} &  2      & 64      & (3,3) \\
			& 2      & 64      & (3,3) \\
			& 2      & 128     & (3,3) \\
			& 2      & 128     & (3,3) \\
			& 2      & 128     & (3,3) \\
			& 2      & 256     & (3,3) \\
			& 2      & 256     & (3,3) \\
			& 2      & 256     & (3,3) \\
			& 2      & 256    & (3,3) \\ \hline
		\end{tabular}
	\end{minipage}
	\begin{minipage}[t]{0.48\linewidth}
		\vspace{1.3cm}
		\centering
		\begin{tabular}{ccc}
			\hline
			& Input & Output  \\ \hline
			\multirow{3}{*}{\rotatebox{90}{MLP}} 
			& 16      & 32       \\
			& 32      & 32     \\
			& 32      & 32     \\
			\hline
		\end{tabular}
		\vspace{0.35cm}
		
		\label{Table: settings in DSOM}
		
	\end{minipage}
\end{table}

\section{Visualization of the output of DSOM on Bicocca dataset.}
\label{Visualization of the output of DSOM on Bicocca dataset.}
\begin{figure}[h]
	\centering
	\subfigure[Scene]{
		\begin{minipage}[t]{0.48\linewidth}
			\centering
			\includegraphics[width=1.4in]{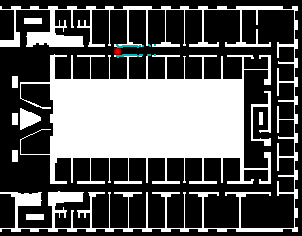}
		\end{minipage}%
	}%
	\centering
	\subfigure[Prediction]{
		\begin{minipage}[t]{0.48\linewidth}
			\centering
			\includegraphics[width=1.4in]{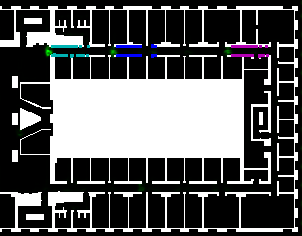}
		\end{minipage}%
	}%
	\vspace{-0.2cm}
	\caption{Visualization results on the output of DSOM on Bicocca dataset.}
	\label{fig: results on Multi-modal Bicocca}
\end{figure}

In Figure \ref{fig: results on Multi-modal Bicocca}, DSOM, which is trained by simulation observation on the blue print map, is able to take the real observation as input and generate a distribution with various numbers of probability peaks including the ground truth.

\vspace{-0.1cm}



\section{Details about AdaM MCL}
\begin{algorithm}[h]
		\label{algorithm PGN Localization Filter}
		\caption{Adaptive Mixture MCL}
		\DontPrintSemicolon
		\KwIn{$\mathcal{P}_{t-1},\ o_t,\ a_{t-1},\ w_{*},\ \mathbf{M}_{env},\ T_{cut}$}
		\KwOut{$\mathcal{P}_{t}$}
		\Begin{
			$\mathbf{PM}\leftarrow$ DSOM$\left(o_t,\mathbf{M}_{env}\right)$; $\mathcal{P}_{t},\ \mathcal{H},\ \mathcal{L}\leftarrow \emptyset$
			\;
			
			\For{$p^i=\left(s^i,w^i_{norm},w^i\right)\in \mathbf{P}_{t-1}$, $T_i=\frac{w^i}{w_*}$}
			{
				\eIf{$T_i>T_{cut}$}
				{
					$\mathcal{H}\leftarrow\mathcal{H}\cup\{p^i\}$
				}
				{
					with probability $T_i$: $\mathcal{H}\leftarrow\mathcal{H}\cup\{p^i\}$ \;
					with probability $(1-T_i)$: $\mathcal{L}\leftarrow\mathcal{L}\cup\{p^i\}$
				}
			}
			\For{$i=1\ to\ \|\mathcal{H}\|$}
			{
				sample $h$ from $\mathcal{H}$ according to $w_{norm}$\;
				sample $h'\sim p\left(h'|a_{t-1},h\right)$ \& $w_{h'}=p\left(o_t|h'\right)$\;
				
				$\mathcal{P}_{t}\leftarrow \mathcal{P}_{t}\cup\left(h',w_{h'}\right)$
			}
			\For{$i=1\ to\ \|\mathcal{L}\|$}
			{
				sample $l'$ according to $\mathbf{PM}$\;
				sample $l\sim p\left(l'|a_{t-1},l\right)$ \& $w_{l'}=w_{l}$\;
				
				$\mathcal{P}_{t}\leftarrow \mathcal{P}_{t}\cup\left(l',w_{l'}\right)$\;
			}
			$w_{norm}\leftarrow$ normalize $w\in \mathcal{P}_{t}$ \& return $\mathcal{P}_{t}$\;
		}
	\end{algorithm}

\vspace{-0.6cm}
\section*{Acknowledgment}
This work was supported in part by the National Nature Science Foundation of China (61903332) and by the Zhejiang Public Welfare Technology Research Program (LGG21F030012).

\ifCLASSOPTIONcaptionsoff
  \newpage
\fi



%
\tiny
\bibliographystyle{IEEEtran}
\bibliography{IEEEabrv,./IEEEexample}

%




\end{document}